\pdfoutput=1

\documentclass[11pt]{article}

\usepackage{acl}

\usepackage{times}
\usepackage{latexsym}

\usepackage[T1]{fontenc}

\usepackage[utf8]{inputenc}

\usepackage{microtype}
\usepackage{bm}
\usepackage{mathtools}
\usepackage{amsmath,amsfonts,amssymb,amsthm}
\usepackage[ruled,noend]{algorithm2e}
\usepackage{booktabs}
\usepackage{multirow}
\usepackage{enumitem}

\usepackage{cleveref}
\crefformat{section}{\S#2#1#3}
\crefformat{subsection}{\S#2#1#3}
\crefformat{subsubsection}{\S#2#1#3}
\crefrangeformat{section}{\S\S#3#1#4 to~#5#2#6}
\crefmultiformat{section}{\S\S#2#1#3}{ and~#2#1#3}{, #2#1#3}{ and~#2#1#3}
\usepackage{refstyle}
\Crefformat{figure}{#2Figure~#1#3}
\Crefmultiformat{figure}{Figures~#2#1#3}{ and~#2#1#3}{, #2#1#3}{ and~#2#1#3}
\Crefformat{table}{#2Table~#1#3}
\Crefmultiformat{table}{Tables~#2#1#3}{ and~#2#1#3}{, #2#1#3}{ and~#2#1#3}
\Crefformat{appendix}{#2Appx.~\S#1#3}
\crefformat{algorithm}{Alg.~#2#1#3}
\Crefformat{equation}{#2Eq.~#1#3}
\newcommand{\stitle}[1]{\vspace{0.3em}\noindent{\bf #1}}

\SetKwInput{KwInput}{Input}
\SetKwInput{KwOutput}{Output}

\author{Wenxuan Zhou$^1$, Fangyu Liu$^2$, Huan Zhang$^3$, Muhao Chen$^1$ \\
$^1$University of Southern California; $^2$University of Cambridge;\\ $^3$Carnegie Mellon University \\
\texttt{\{zhouwenx, muhaoche\}@usc.edu}; \texttt{fl399@cam.ac.uk};\\ \texttt{huan@huan-zhang.com}
}

\title{Sharpness-Aware Minimization with Dynamic Reweighting}

\begin{document}
\maketitle
\begin{abstract}
Deep neural networks are often overparameterized and may not easily achieve model generalization.
Adversarial training has shown effectiveness in improving generalization by regularizing the change of loss on top of adversarially chosen perturbations.
The recently proposed sharpness-aware minimization~(SAM) algorithm conducts adversarial weight perturbation, encouraging the model to converge to a flat minima.
SAM finds a \emph{common} adversarial weight perturbation \emph{per-batch}. Although \emph{per-instance} adversarial weight perturbations are stronger adversaries and can potentially lead to better generalization performance, their computational cost is very high and thus it is impossible to use per-instance perturbations efficiently in SAM.
In this paper, we tackle this efficiency bottleneck and propose sharpness-aware minimization with dynamic reweighting~($\delta$-SAM). 
Our theoretical analysis motivates that it is possible to approach the stronger, per-instance adversarial weight perturbations using reweighted per-batch weight perturbations. 
$\delta$-SAM dynamically reweights perturbation within each batch according to the theoretically principled weighting factors, serving as a good approximation to per-instance perturbation.
Experiments on various natural language understanding tasks demonstrate the effectiveness of $\delta$-SAM.

\end{abstract}

\section{Introduction}
Although deep neural networks~(DNNs) have demonstrated promising results in various fields such as natural language understanding~\cite{devlin-etal-2019-bert} and computer vision~\cite{krizhevsky2012imagenet}, they are often overparameterized and can easily overfit the training data~\cite{zhang2021understanding}.
Adversarial training has been proven effective in improving both model generalization~\cite{zhu2019freelb,zhang2020geometry} and adversarial robustness~\cite{madry2018towards,zhang2019theoretically}.
A general approach for adversarial training has been first to augment the inputs with small perturbations that lead to the maximum possible change of loss, and then optimize the model parameters to the direction where the changed amount is minimized.

\begin{figure}
    \centering
    \includegraphics[width=0.99\linewidth]{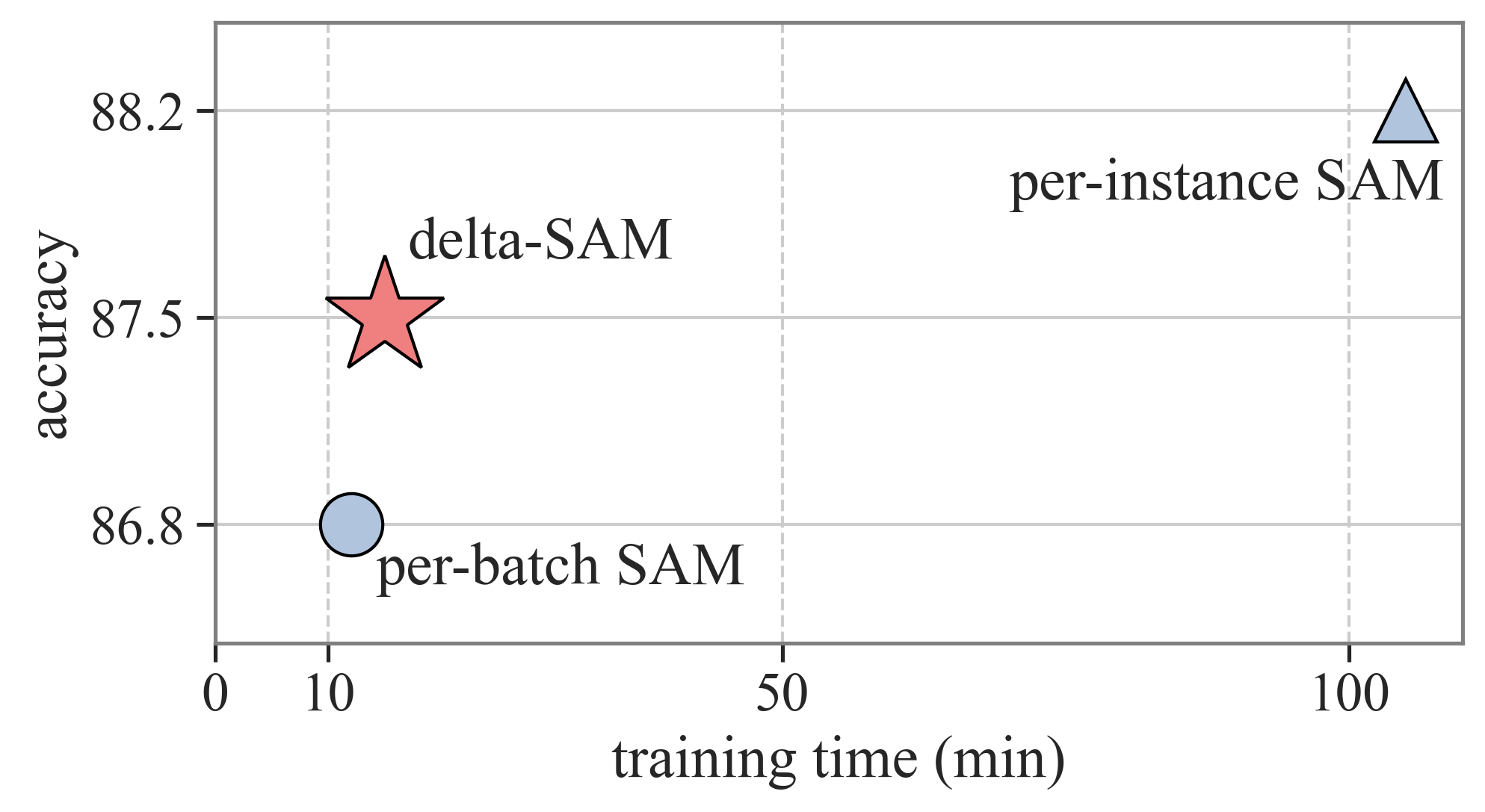}
    \caption{$\delta$-SAM performs close to the computation-intensive per-instance weight perturbation while adding only marginal computation overhead to the standard per-batch SAM. Results shown are from the MRPC dataset. See~\Cref{ssec:analysis} for more detailed results.}
    \label{fig:front_fig}
\end{figure}

Besides perturbing inputs, a recent work of sharpness-aware minimization~(SAM; \citealt{foret2020sharpness}) has further proposed to adversarially perturb model weights.
Such a method works by first adversarially calculating a weight perturbation that maximizes the empirical risk and then minimizing the empirical risk on the perturbed model.
This method demonstrates improved model generalizations across different datasets and models.
In principle, each instance in a batch has its own worst-case weight perturbation and the weight perturbations of different instances need to be calculated separately and cannot be done in a single forward/backward pass.
This leads to a significant increase in computational and memory cost.
To allow a feasible algorithm, SAM approximates \emph{per-instance} perturbations by a single \emph{per-batch} perturbation, where the weight perturbation is calculated on the averaged loss of the batch and shared by all instances in the batch.
However, as the per-batch perturbation represents the average of perturbations yielded by different instances, it is a weaker adversary compared to per-instance perturbations, and may hinder the effectiveness of SAM.

In this paper, we study how to efficiently approximate per-instance weight perturbation for sharpness-aware minimization, while maintaining a similar computational cost to per-batch perturbation.
We first theoretically analyze the gradient posed by the optimization of per-instance perturbation, and find that it can be effectively approximated with a \textit{weighted-batch} perturbation under some assumptions, where the instances with a larger rate of gradient change are up-weighted.
Based on this motivation, we propose sharpness-aware minimization with dynamic reweighting~($\delta$-SAM).
Specifically, we first estimate the Hessian and gradient norm of each instance by perturbing the loss with a random Gaussian noise on model weights.
Next, $\delta$-SAM dynamically reweights the loss within each batch of training instances, and then calculate a shared weight perturbation that maximizes the reweighted batch loss.
Finally, we update the perturbed model on the original~(unweighted) batch.
Compared to SAM, $\delta$-SAM only requires extra computation cost in estimation of the rate of gradient change, which can be efficiently performed using three additional forward passes.

We evaluate $\delta$-SAM on finetuning pretrained language models~(PLMs).
Experiments on standard GLUE benchmark \cite{wang-etal-2018-glue}, self-supervised Semantic Textual Similarity (STS), and abstractive summarization tasks show that besides significantly outperforming 
base models, $\delta$-SAM also consistently outperforms SAM with only $18\%$ extra computational cost in average.

The main contributions of this paper are three-fold. First, we analyze the training objective of per-instance weight perturbation and find that under some assumptions, it can be approximated by a weighted-batch perturbation, where instances are efficiently reweighted according to their caused rates of gradient changes.
Second, we propose to use random perturbations as estimations to efficiently realize the weighting scheme.
Third, we evaluate $\delta$-SAM on a diverse set of datasets and find consistent improvements across the board. 

\section{Related Work}
\stitle{Model Generalization.}
Deep neural networks are often overparameterized and may suffer from poor generalization~\cite{zhang2021understanding}.
A lot of efforts have been devoted to improve the generalization of neural models, leading to methods including data augmentation~\cite{sennrich-etal-2016-improving,wei-zou-2019-eda,sun-etal-2020-mixup,kumar-etal-2020-data,thakur-etal-2021-augmented}, regularization~\cite{loshchilov2017decoupled,xuhong2018explicit,liang2021r}, and improved optimization processes~\cite{izmailov2018averaging,mobahi2020self,heo2020adamp}.
These methods consider different aspects of generalization and may be combined to achieve better performance.
Among them, adversarial training~\cite{goodfellow2014explaining} has demonstrated its effectiveness in improving model generalization without the need of any extra data or external knowledge, 
and has been widely attempted to enhance NLP models~\cite{zhu2019freelb,jiang-etal-2020-smart,pereira-etal-2021-targeted,li2021token}.
Adversarial training works by adversarially perturbing the input embedding and either minimize the adversarial risk or regularizes the change of risk to be small.
Specifically, FreeLB~\cite{zhu2019freelb} uses projected gradient descent~(PGD; \citealt{madry2018towards}) to generate adversarial perturbations on input embedding, and recycles the computed gradients when updating model parameters in adjacent steps~\cite{shafahi2019adversarial} to reduce the computational costs.
TAT~\cite{pereira-etal-2021-targeted} improves FreeLB by prioritizing the most frequently mispredicted classes in perturbation calculation.
TAVAT~\cite{li2021token} uses a token-level accumulated perturbation vocabulary to guide the initialization in PGD.
However, these works only consider the robustness on input feature representation, while we consider the robustness of all model weights.

\stitle{Sharpness-Aware Minimization.}
\citet{foret2020sharpness} leverage the correlation between flat minima and better model generalization and propose SAM for training deep neural models that are robust to adversarial weight perturbations.
It has demonstrated effectiveness in tasks on both vision~\cite{chen2021vision,zheng2021regularizing} and language~\cite{bahri2021sharpness} modalities.
Several variants of SAM have been proposed to improve its efficiency or effectiveness.
For efficiency, \citet{brock2021high} propose to speed up SAM by perturbing fewer instances in the batch; \citet{du2021efficient} introduce stochastic weight perturbation and sharpness-sensitive data selection to reduce the computational overhead.
For effectiveness; \citet{kwon2021asam} propose to adaptively set the SAM's radius such that it is invariant to parameters' scales; \citet{zhuang2021surrogate} introduce a gradient ascent optimization step in the perturbed model's orthogonal direction to achieve better flatness.
In contrast to the aforementioned studies, this paper, for the first time, tackles how to narrow the gap of per-instance and per-batch weight perturbation.
Accordingly, we propose an efficient approximation to per-instance weight perturbation, which shows improved results on several NLP tasks while does not bring much computational overheads.

\begin{algorithm}[!t]
    \caption{SAM and $\delta$-SAM}\label{algo::main}
    \textbf{Input:} model $f_{\bm{w}}$, training set $\mathcal{S} \triangleq \{(\bm{x}_i, \bm{y}_i)\}_{i=1}^{|\mathcal{S}|}$, loss function $l: \mathcal{W}\times \mathcal{X}\times \mathcal{Y}\rightarrow \mathbb{R}^+$, batch size $N$, neighborhood size $\rho\in \mathbb{R}^+$, optimizer $h$. \\
    \textbf{Output:} a flat solution $\hat{\bm{w}}$.\\
    Initialize model weights $\bm{w}$. \\
    \While {not converge}{
    Sample a batch $\mathcal{B}=\{(\bm{x}_j, \bm{y}_j)\}_{j=1}^N$. \\
    \emph{\textcolor{teal}{For $\delta$-SAM:}} \\
    \emph{\textcolor{teal}{Reweigh $\mathcal{B}$ by~\Cref{eq:es1}.}}\\
    Compute gradient $\nabla l_{\mathcal{B}}(\bm{w})$ of the (reweighted) batch's empirical risk. \\
    Perturb the model weights by $\bm{\epsilon}^*=\rho \nabla l_{\mathcal{B}}(\bm{w}) / \left\lVert\nabla l_{\mathcal{B}}(\bm{w})\right\rVert_2$. \\
    Update $\bm{w}$ w.r.t. the unweighted empirical risk $\frac{1}{N}\sum_{j=1}^N l_j(\bm{w}+\bm{\epsilon}^*)$ with the optimizer $h$. \\
    }
\end{algorithm}

\section{Sharpness-Aware Minimization~(SAM)}
\label{sec:sam}
In this section, we briefly review the principle of SAM and discuss its limitations.

Literature has observed a direct correlation between flat minima and better model generalization, both empirically and theoretically~\cite{keskar2016large,dziugaite2017computing,li2018visualizing,jiang2019fantastic}.
To find a flat loss landscape, SAM~\cite{foret2020sharpness} adversarially perturbs the model weights and optimizes the following min-max objective on a batch of size $N$:
\begin{equation}
\label{eq:per_batch}
    \min_{\bm{w}} \max_{\bm{\epsilon}: \left\lVert\bm{\epsilon}\right\rVert_2\le \rho}\frac{1}{N}\sum_{i=1}^N l_i(\bm{w}+\bm{\epsilon}),
\end{equation}
where given the model weights $\bm{w}$, the inner maximization seeks for a perturbation $\bm{\epsilon}$ with $L_2$-norm $\le \rho$ that maximizes the empirical risk, and the outer minimization minimizes the empirical risk of the perturbed model.
This training objective aims at finding model parameters whose neighborhood has a uniformly low training loss.
As finding the exact solution to $\bm{\epsilon}$ is NP-hard, SAM estimates the solution $\bm{\epsilon}^*$ of the inner maximization with a single-step gradient descent on the empirical risk of the batch:
\begin{align*}
    l(\bm{w}) &= \frac{1}{N}\sum_{i=1}^N l_i(\bm{w}), \\
    \bm{\epsilon}^*&\approx\arg\!\max_{\bm{\epsilon}: \left\lVert\bm{\epsilon}\right\rVert_2\le \rho} l(\bm{w}) + \bm{\epsilon}^\intercal \nabla l(\bm{w}) \\
    &= \rho \nabla l(\bm{w}) / \left\lVert\nabla l(\bm{w})\right\rVert_2.
\end{align*}
The outer minimization can be performed with a standalone optimizer~(e.g., Adam; \citealt{kingma2015adam}).
SAM roughly doubles the computational cost of training the model, requiring two forward and two backward passes for each batch.
The SAM algorithm is outlined in~\Cref{algo::main}.

Besides perturbing by batches, weight perturbation can also be performed on individual instances:
\begin{equation}
\label{eq:per_instance}
    \min_{\bm{w}} \frac{1}{N}\sum_{i=1}^N \max_{\bm{\epsilon}_i: \left\lVert\bm{\epsilon}_i\right\rVert_2\le \rho} l_i(\bm{w}+\bm{\epsilon}_i),
\end{equation}
where $\bm{\epsilon}_i$ is calculated by single-step gradient descent on individual instances.
This approach is similar to many adversarial training methods in NLP, such as VAT~\cite{miyato2018virtual} and FreeLB~\cite{zhu2019freelb}, except that the perturbation is computed on model weights instead of input embedding only.
We refer to the objectives of~\Cref{eq:per_batch} and~\Cref{eq:per_instance} as \emph{per-batch weight perturbation} and \emph{per-instance weight perturbation}, respectively.
It is noted in the same paper by \citet{foret2020sharpness} that per-instance weight perturbation produces a smaller test error and is a better predictor of model generalization.

Despite its effectiveness, per-instance weight perturbation increases the computational and memory cost significantly, requiring $2N$ forward and $2N$ backward passes for a batch of size $N$.
Because per-instance weight perturbation modifies all model weights independently, the perturbation for each individual instance needs to be computed on a distinct model copy.
Therefore, per-instance weight perturbation can be computationally unaffordable for large-scale training.

\section{SAM with Dynamic Reweighting}
In this paper, we seek to improve SAM with a better adversary on weight perturbations. 
As the per-batch weight perturbation adopted by SAM weakens the adversarial training, we propose a simple yet effective modification of SAM, $\delta$-SAM~(SAM with dynamic reweighting), that can approximate per-instance weight perturbation without requiring much additional computational cost.
Our reweighting approach is motivated by a theoretical analysis on approximating the per-instance weight perturbation to justify its superior efficiency. Based on this motivation, we then illustrate how $\delta$-SAM is realized in implementation.

\subsection{Theoretical Motivations}
\label{sec:theoretical_motivation}
In this subsection, we motivate our dynamic reweighting approach by formally analyzing the training objective posed by per-instance perturbation, and show that it can be approximated with a weighted-batch perturbation, which motivates our $\delta$-SAM algorithm.

\stitle{Preliminary.}
We motivate our approach from the perspective of sharpness in SAM, which quantifies the flatness of loss landscape as the \emph{increase of loss} in the neighborhood region of model weights.
The sharpness of per-batch and per-instance weight perturbations are defined as:
\begin{align*}
    \mathcal{R}_\text{batch}&=\max_{\bm{\epsilon}: \left\lVert\bm{\epsilon}\right\rVert_2\le \rho}\frac{1}{N}\sum_{i=1}^N \left(l_i(\bm{w} + \bm{\epsilon}) - l_i(\bm{w})\right), \\
    \mathcal{R}_\text{inst}&=\frac{1}{N}\sum_{i=1}^N \max_{\bm{\epsilon}_i: \left\lVert\bm{\epsilon}_i\right\rVert_2\le \rho} \left(l_i(\bm{w} + \bm{\epsilon}_i) - l_i(\bm{w})\right).
\end{align*}
Due to non-shared $\bm{\epsilon}_i$, $\mathcal{R}_\text{inst} \ge \mathcal{R}_\text{batch}$, suggesting stronger regularization effects of $\mathcal{R}_\text{inst}$. However, $\mathcal{R}_\text{inst}$ is expensive to compute, since $\epsilon_i$ in the $N$ inner maximization problems must be calculated by gradient descent on $N$ individual instances, for which $O(N)$ backward passes through the network are needed. In the analysis below, we show how to approximate the stronger $\mathcal{R}_\text{inst}$ with a weighted per-batch weight perturbation.

We start by considering the second-order expansion of a general empirical risk $l_i$ for instance $i$:
\begin{equation*}
    l_i(\bm{w}+\bm{\epsilon})=l_i(\bm{w}) + \nabla l_i(\bm{w})^\intercal \bm{\epsilon} + \frac{1}{2} \bm{\epsilon}^\intercal \bm{H}_i(\bm{w}) \bm{\epsilon}.
\end{equation*}
To allow a tractable theoretical analysis, we assume that the Hessian is a low-rank, positive definite matrix $\bm{H}_i(\bm{w}) =a_i \nabla l_i(\bm{w}) \nabla l_i(\bm{w})^\intercal, (a_i>0)$.
Then, we obtain the perturbations in $\mathcal{R}_\text{batch}$ and $\mathcal{R}_\text{inst}$ under second-order approximation in closed-form with one-step gradient descent, to align with the practice of SAM:
\begin{align*}
    \bm{\epsilon} &= \rho \nabla l(\bm{w}) / \left\lVert \nabla l(\bm{w})\right\rVert_2, \\
    \bm{\epsilon}_i &= \rho \nabla l_i(\bm{w}) / \left\lVert \nabla l_i(\bm{w})\right\rVert_2,
\end{align*}
where $\nabla l(\bm{w})=\frac{1}{N}\sum_{i=1}^N \nabla l_i(\bm{w})$ is the average gradient of the batch.

\stitle{Training Objective.} 
After $\mathcal{R}_\text{batch}$ or $\mathcal{R}_\text{inst}$ is obtained, SAM will compute $\frac{\partial}{\partial \bm{w}} \mathcal{R}_\text{batch}$ or $ \frac{\partial}{\partial \bm{w}}\mathcal{R}_\text{inst}$ and update model weights to minimize the loss. To aim for a more effective perturbation, we seek to align with the gradient $\mathcal{R}_\text{inst}$, which determines how model weights will be updated under the strong per-instance adversary. We hope to update the model weights in a ``similar'' manner as the per-instance adversary, while not explicitly computing the expensive term $\mathcal{R}_\text{inst}$. Here ``similar'' means the cosine similarity between the gradient of $\mathcal{R}_\text{inst}$ and our new objective is positive.\footnote{If the cosine similarity is positive, then optimizing $\mathcal{R}$ also leads to optimized $\mathcal{R}_\text{inst}$.}

We thereby first derive the gradient of per-instance weight perturbation. 
Specifically, after calculating $\bm{\epsilon}_i$ in the inner maximization step ($\bm{\epsilon}_i$ is not differentiated in outer minimization), the gradient of per-instance perturbation is:
\begin{align}
    \frac{\partial}{\partial \bm{w}} \mathcal{R}_\text{inst}&=\frac{1}{N} \sum_{i=1}^N \left(\nabla l_i(\bm{w} + \bm{\epsilon}_i) - \nabla l_i(\bm{w})\right) \nonumber \\
    &= \frac{1}{N}\sum_{i=1}^N \bm{H}_i (\bm{w}) \bm{\epsilon}_i \nonumber \\
    &= \frac{1}{N}\sum_{i=1}^N \rho a_i \left\lVert \nabla l_i(\bm{w})\right\rVert_2 \nabla l_i(\bm{w}). \label{eq:grad_rinst}
\end{align}


In this paper, we aim at finding a per-batch perturbation $\bm{\epsilon}'$ that produces the gradient whose direction is aligned with the per-instance gradient, so model weights will be updated similarly using gradient based optimizers. 
Specifically, for a shared perturbation $\bm{\epsilon}^\prime$ and $\mathcal{R}$ defined as  $\mathcal{R}:=\frac{1}{N}\sum_{i=1}^N \left(l_i(\bm{w} + \bm{\epsilon}') - l_i(\bm{w})\right)$, its gradient is:
\begin{align*}
    \frac{\partial}{\partial \bm{w}} \mathcal{R}&=\frac{1}{N} \sum_{i=1}^N \left(\nabla l_i(\bm{w} + \bm{\epsilon}') - \nabla l_i(\bm{w})\right)\\
    &= \left ( \frac{1}{N}\sum_{i=1}^N \bm{H}_i(\bm{w}) \right ) \bm{\epsilon}'.
\end{align*}
Compared to the ordinary $\mathcal{R}_\text{batch}$, here we propose to use a different $\bm{\epsilon}^\prime$.
Our goal is that optimizing $\mathcal{R}$ also leads to smaller $\mathcal{R}_\text{inst}$; that is, we seek to find an $\bm{\epsilon}'$ such that $\left ( \frac{\partial}{\partial \bm{w}}\mathcal{R} \right )^\intercal \frac{\partial}{\partial \bm{w}} \mathcal{R}_\text{inst} > 0$.
An easy choice would be:
\begin{equation}
\label{eq:epsilon_prime}
\bm{\epsilon}'=\rho \cdot \left ( \frac{\partial}{\partial \bm{w}} \mathcal{R}_\text{inst} \right ) \Big / \left\lVert \frac{\partial}{\partial \bm{w}} \mathcal{R}_\text{inst}\right\rVert_2 .
\end{equation}
Because each $\bm{H}_i$ is positive definite under our assumptions, $\frac{1}{N}\sum_{i=1}^N \bm{H}_i(\bm{w})$ is also positive definite, then we have:
\begin{align*}
    &\;\left ( \frac{\partial}{\partial \bm{w}}\mathcal{R} \right )^\intercal \frac{\partial}{\partial \bm{w}} \mathcal{R}_\text{inst}
    \\& \propto \left ( \frac{\partial}{\partial \bm{w}} \mathcal{R}_\text{inst} \right )^\intercal \left(\frac{1}{N}\sum_{i=1}^N \bm{H}_i(\bm{w})\right) \left ( \frac{\partial}{\partial \bm{w}} \mathcal{R}_\text{inst} \right )
    \\& > 0.
\end{align*}
Thus, under this specific $\bm{\epsilon}^\prime=\frac{\partial}{\partial \bm{w}} \mathcal{R}_\text{inst}$, $\frac{\partial}{\partial \bm{w}}\mathcal{R}$ is aligned with $\frac{\partial}{\partial \bm{w}} \mathcal{R}_\text{inst}$.
Although $\frac{\partial}{\partial \bm{w}} \mathcal{R}_\text{inst}$ is the gradient that we aim to approximate and not computable here, the key observation of~\Cref{eq:epsilon_prime} is that \textit{per-instance weight perturbation can be optimized by using a perturbation shared by all instances in the batch}.
Therefore, we attempt to perturb the model with only a (rough) estimation of $\bm{\epsilon}^\prime$.
Now the next challenge is how to efficiently derive such estimation.

An important observation is that under our assumptions on $\bm{H}_i(\bm{w})$ and the use of second order approximations, $\frac{\partial}{\partial \bm{w}} \mathcal{R}_\text{inst}$ can be calculated by using one time backpropagation on a \emph{reweighted batch}. To see this fact, we define weights $g_i=a_i \left\lVert \nabla l_i(\bm{w})\right\rVert_2$ and the reweighted batch is:
\[
l_\text{reweighted} = \frac{1}{N}\sum_{i=1}^N g_i l_i(\bm{w}) .
\]
Then, treating $g_i$ as constants, $\frac{\partial}{\partial \bm{w}} l_\text{reweighted} \propto \frac{\partial}{\partial \bm{w}} \mathcal{R}_\text{inst}$, as defined in Eq.~\ref{eq:grad_rinst}.
Compared to per-batch SAM, this reweighting only requires small extra computation cost on calculating the instance weights $g_i$. We introduce how to estimate these weights efficiently in the following section.

\subsection{Implementation}
In $\delta$-SAM, the key problem in implementation is how to efficiently estimate the instance weight $g_i$.
We solve this problem by sampling random perturbations.
Specifically, for random perturbation $\bm{r}$, where $\bm{r}_i$ follows Gaussian distribution $\mathcal{N}(0, \sigma \bm{I})$, under the same assumptions as in Section~\ref{sec:theoretical_motivation}, we have:
\begin{align*}
    &\;E[\left(l_i(\bm{w} + \bm{r}) - l_i(\bm{w} - \bm{r})\right)^2]
    \\ &=E[\nabla l_i(\bm{w})^\intercal \bm{r} \cdot \nabla l_i(\bm{w})^\intercal \bm{r}]\\
    &= \sigma^2 \left\lVert \nabla l_i(\bm{w})\right\rVert_2^2.\\
    &\;E[l_i(\bm{w} + \bm{r}) + l_i(\bm{w} - \bm{r}) - 2l_i(\bm{w})]
    \\ &=E[\bm{r}^\intercal \bm{H_i} \bm{r}]\\
    &= a_i\sigma^2 \left\lVert \nabla l_i(\bm{w})\right\rVert_2^2.
\end{align*}
Therefore, by sampling random perturbations and take the expectation, we can get unbiased estimations of $a_i$ and the gradient norm $\left\lVert \nabla l_i(\bm{w})\right\rVert_2$.
Each estimation takes three forward passes for calculating $l_i(\bm{w})$, $l_i(\bm{w} + \bm{r})$, and $l_i(\bm{w} - \bm{r})$.
As we do not need to save the intermediate states for back-propagation (\texttt{no\_grad} in PyTorch), these forward passes are faster than normal ones.
In $\delta$-SAM, for the efficiency of the algorithm, we only sample one (shared) $\bm{r}\sim\mathcal{N}(0, \sigma \bm{I})$ for each batch in $\delta$-SAM, and then calculate the instance weight $g_i$ by:
\begin{equation}
    g_i=\frac{\left|l_i(\bm{w} + \bm{r}) + l_i(\bm{w} - \bm{r}) - 2 l_i(\bm{w})\right|}{\max(\left|l_i(\bm{w} + \bm{r}) - l_i(\bm{w} - \bm{r})\right|, \eta)},\label{eq:es1}
\end{equation}
where $\eta$ is a hyperparameter for avoiding division by zero.
After deriving the instance weights, the \textit{weighted-batch} weight perturbation can be computed by:
\begin{align}
    \nabla l_\mathcal{B}(\bm{w}) &= \nabla \left(\sum_i^N g_i l_i(\bm{w})\right), \label{eq:weighted_pert}\\
    \bm{\epsilon}^*&= \rho \nabla l_\mathcal{B}(\bm{w}) / \left\lVert \nabla l_\mathcal{B}(\bm{w}) \right\rVert_2. \label{eq:renorm}
\end{align}


We hereby summarize our algorithm, as outlined in~\Cref{algo::main}.
Modifications made for $\delta$-SAM are highlighted in blue.
Given a batch $\mathcal{B}$, we first dynamically reweigh the instances by \Cref{eq:es1}, then estimate the perturbation $\bm{\epsilon}^*$ that maximizes the reweighted loss by a single-step gradient descent as shown in~\Cref{eq:weighted_pert} and~\Cref{eq:renorm}, and finally minimize the empirical risk of the perturbed model on the original~(unweighted) batch.

\begin{table*}[!t]
\centering
\small
    \begin{tabular}{lccccccccc}
    \toprule
    Model&  avg.& MNLI& QQP& RTE& QNLI& MRPC& CoLA& SST2& STS-B \\
    &&Acc-m& Acc& Acc& Acc& Acc& Mcc& Acc& Pearson\\
    \midrule
    BERT$_\textsc{base}$& 82.9& 83.8& 91.0& 68.2& 90.8& 85.3& 62.3& 92.4& 89.3\\
    R-Drop~\cite{liang2021r}& 84.1& \textbf{85.5}& 91.4& 71.1& \textbf{92.0}& 87.3& 62.6& 93.0& 89.6\\
    \midrule
    SAM& 83.9& 85.0& 91.6& 69.3& 91.7& 88.2& 63.1& 93.0& 89.4\\
    $\delta$-SAM& \textbf{84.7}& 85.2& \textbf{91.7}& \textbf{72.2}& 91.5& \textbf{89.5}& \textbf{63.8}& \textbf{93.7}& \textbf{89.7}\\
    \midrule
    \midrule
    RoBERTa$_\textsc{large}$~\cite{liu2019roberta}& 88.9& 90.2& 92.2& 86.6& 94.7& 90.9& 68.0& 96.4& 92.4\\
    R-Drop~\cite{liang2021r}& 89.7& 90.9& 92.5& 88.4& 95.2& 91.4& 70.0& 96.9& 92.5 \\
    FreeLB~\cite{zhu2019freelb}& 89.8& 90.6& \textbf{92.6}& 88.1& 95.0& 91.4& 71.1& 96.7& \textbf{92.7} \\
    SMART~\cite{jiang-etal-2020-smart}& -& \textbf{91.1}& 92.4& \textcolor{gray}{92.0\rlap{$^*$}}& \textbf{95.6}& \textcolor{gray}{89.2\rlap{$^*$}}& 70.6 & 96.9& \textcolor{gray}{92.8\rlap{$^*$}} \\
    R3F~\cite{aghajanyan2020better}& -&\textbf{91.1}& 92.4& 88.5& 95.3& 91.6&  \textbf{71.2}&\textbf{97.0}& -\\
    \midrule
    SAM& 89.6&91.0& 92.3&88.5& 95.0&91.4& 69.2& 96.7& 92.4 \\
    $\delta$-SAM& \textbf{90.1}&\textbf{91.1}& 92.5&\textbf{89.2}& 95.1&\textbf{92.2}& 71.1& 96.9& \textbf{92.7} \\
    \bottomrule
    \end{tabular}
    \caption{Results on the development set of the GLUE benchmark. $^*$ denotes results derived from the model intermediately trained on the MNLI dataset (not comparable to other results), while others are derived by finetuning the original BERT/RoBERTa. The results of BERT$_\textsc{base}$ are from the reimplementation by~\citet{liang2021r}.}\label{tab::main_result}
\end{table*}

\section{Experiments}
\label{sec:exp}



This section presents experimental evaluation of $\delta$-SAM based on the GLUE benchmark tasks (\Cref{ssec:glue}), self-supervised Semantic Textual Similarity (STS) tasks (\Cref{ssec:sts}), and the CNN/DailyMail abstractive summarization task (\Cref{ssec:summ}).
We also provide additional analyses to further illustrate the performance of $\delta$-SAM (\Cref{ssec:analysis}).

\subsection{Baseline Methods}

We compare SAM and $\delta$-SAM to the following baseline methods, which were all proposed for improving the generalization of PLMs:

\begin{itemize}[leftmargin=1em]
\setlength\itemsep{0em}
\item \textbf{R-Drop}~\cite{liang2021r} enforces the prediction of the same instances augmented by different dropout masks to be similar with a consistency term~(KL divergence for classification and mean squared error for regression), which leads to improved performance on various language and vision tasks.
\item \textbf{R3F}~\cite{aghajanyan2020better} also uses a consistency term to make the prediction of the same instance to be similar.
Besides augmenting the instances by different dropout masks, it further adds random uniform or normal noise to input embedding in PLMs.
Therefore, R3F can be regarded as an extension to R-Drop.
\item \textbf{FreeLB}~\cite{zhu2019freelb} adversarially perturbs the token embedding using a multi-step projected gradient descent~(PGD; \citealt{madry2018towards}) to maximize the empirical risk and regularizes the adversarial risk to be small.
\item \textbf{SMART}~\cite{jiang-etal-2020-smart} is a framework that combines multiple techniques for improving model generalization, including adversarial training, and improved optimizer and regularization techniques.
In terms of adversarial training, it 
perturbs the input embedding with PGD to maximize the empirical risk.
It then uses a consistency term to regularize the change of risk to be small, for which the consistency term is defined as the KL divergence for classification and mean squared loss for regression.
\end{itemize}

\begin{table*}[!t] 
\centering
\small
\begin{tabular}{l ccccccccccc}
\toprule
Model  & avg. & STS12 & STS13 & STS14 & STS15 & STS16 & STS-B & SICK-R\\
\midrule
Mirror-BERT$_\textsc{base}$ &  74.85 & 67.87 & 80.98 & 71.84 & 81.58 & 74.41 & 77.78 & 69.53  \\
+ FreeLB& 75.71& 69.78& 80.81& 72.87& 82.55& 74.77& 79.01& \textbf{70.40}\\
+ R3F & 75.27 & 68.58 & 80.93& 72.51& 81.97& 75.47& 77.62& 69.79  \\
+ SAM& 75.50& 68.37& 82.16& 72.88& 82.22& 74.71& 77.98& 70.20 \\
+ $\delta$-SAM & 75.72 & 68.44 & 82.30 & 73.12 & 82.27 & 75.12 & 78.83 & 69.94  \\
+ SAM w/ random noise& 76.44& \textbf{70.09}& 83.53& 74.22& 82.67& 77.61& \textbf{80.16}& 66.81 \\
+ $\delta$-SAM w/ random noise& \textbf{76.71}& 69.73& \textbf{83.73}& \textbf{74.58}& \textbf{83.01}& \textbf{77.72}& 79.08& 69.11 \\
\midrule
\midrule
Mirror-RoBERTa$_\textsc{base}$& 74.98 & 64.49 & 81.69 & 73.32 & 79.78 & 77.49 & 78.53 & 69.56  \\
+ FreeLB& 75.73& 65.52& 82.41& 74.00& 80.72& 78.60& 79.06& 69.81 \\
+ R3F& 75.32  & 64.58 & 82.12 & 73.62 & 80.27 & 77.90 & 78.66& 70.13 \\
+ SAM& 75.18& 64.86& 81.96& 73.56& 79.82& 77.06& 78.83& 70.21 \\
+ $\delta$-SAM & 75.27 & 64.97 & 81.89 & 73.46 & 80.12 & 77.62 & 78.96 & 69.89 \\
+ SAM w/ random noise& 75.90& 66.65& 82.52& 74.10& 80.81& 78.47& 79.02& 69.71 \\
+ $\delta$-SAM w/ random noise& \textbf{76.31}& \textbf{66.94}& \textbf{82.86}& \textbf{74.46}& \textbf{81.13}& \textbf{78.91}& \textbf{79.52}& \textbf{70.32} \\
\bottomrule
\end{tabular}
\caption{Results on self-supervised STS tasks. For all datasets, we report the average Spearman's $\rho$ of 5 runs of training using 5 fixed random seeds.}
\label{tab:unsupervised_sts_full}
\end{table*}


\subsection{GLUE Tasks}\label{ssec:glue}
\stitle{Task Setup.}
We first evaluate $\delta$-SAM on the GLUE benchmark~\cite{wang-etal-2018-glue}.
In this experiment,
we use both BERT$_\textsc{base}$ and RoBERTa$_\textsc{large}$ as the encoders.
To ensure a fair comparison, for task-specific hyperparameters including batch sizes, optimizers, learning rates, training steps, weight decay, dropout rates, and learning rate scheduling, we strictly replicate the values from R-Drop~\cite{liang2021r}.
For SAM and $\delta$-SAM, we search $\rho$ in $\{0.01, 0.02, 0.05\}$ and $\eta$ in $\{\texttt{1e-4}, \texttt{2e-4}, \texttt{5e-4}, \texttt{1e-3}\}$.
For $\sigma$ in random perturbations, we find that rescaling the random Gaussian perturbation to an $L_2$-norm of $\rho$ achieves promising results, so we simply set $\sigma=1$ and rescale the random perturbation afterwards.
Following the evaluation settings of R3F and R-drop, we report the best result on the development set out of 5 runs of training with different random seeds.

\stitle{Results and Discussion.}
Results are shown in \Cref{tab::main_result}.
We observe that in average, SAM improves BERT$_\textsc{base}$ and RoBERTa$_\textsc{large}$ by $1.0\%$ and $0.7\%$, respectively, showing that SAM improves the generalization of PLMs, being consistent with the findings in the recent work~\cite{bahri2021sharpness}.
However, its performance is still worse than other compared methods.
On the other hand, $\delta$-SAM improves BERT$_\textsc{base}$ and RoBERTa$_\textsc{large}$ by $1.8\%$ and $1.2\%$, respectively, and also achieves better or comparable results compared to other methods, demonstrating its effectiveness.
In terms of individual tasks, the performance gain of $\delta$-SAM to SAM is larger on smaller datasets~(e.g., MRPC, RTE, CoLA, SST2), while it becomes less prominent on larger datasets.
We hypothesize that due to increased training steps and number of instances in large datasets, the gap between per-batch and per-instance perturbation becomes smaller.
It is also possible that smaller datasets need better generalization so $\delta$-SAM helps more.
Besides, we observe that the improved performance and generalization by $\delta$-SAM is obtained at a merely little average extra computational cost of $18\%$ to SAM (see~\Cref{tab:time} in Appendix for the running time of models).
Taking BERT$_\textsc{base}$ and the SST2 dataset as an example, the average running time is 118/132 min for SAM/$\delta$-SAM, respectively, meaning that $\delta$-SAM is only $12\%$ slower than SAM to approximate per-instance perturbation.


\subsection{Self-supervised STS}\label{ssec:sts}
\stitle{Model.}
To conduct self-supervised STS evaluation,
we apply $\delta$-SAM to the training process of Mirror-BERT$_\textsc{base}$ and Mirror-RoBERTa$_\textsc{base}$~\cite{liu-etal-2021-fast}, which are SOTA self-supervised sentence embedding frameworks.
Similar to SimCSE~\cite{gao-etal-2021-simcse}, Mirror-BERT embeds a sentence $x$ with the same encoder but different dropout masks to get two sentence embedding $h_1$ and $h_2$, and optimizes $h_1$ and $h_2$ to be similar using contrastive loss.
This training objective resembles R-Drop and SMART.
Empirically, we find that applying adversarial training~(FreeLB, SAM, and $\delta$-SAM) to only one embedding~(e.g. $h_2$ only) achieves much better results than to the contrastive loss, and we use that strategy in experiments.

\stitle{Task Setup.}
We experiment with self-supervised sentence embedding learning on 7 STS datasets including SemEval 2012-2016 datasets (STS12-16, \citealt{agirre2012semeval,agirre2013sem,agirre2014semeval,agirre2015semeval,agirre2016semeval}), STS Benchmark (STS-B, \citealt{cer2017semeval}), and SICK-Relatedness (SICK-R, \citealt{marelli2014sick}).
We strictly follow and replicate the model and experimental setup of Mirror-BERT.
As the training objective of Mirror-BERT is similar to R-Drop and SMART, we compare with other baselines including R3F\footnote{Applying R3F to Mirror-BERT is simply adding random noise to input embedding.} and FreeLB on this task.
Furthermore, as FreeLB and R3F both use random noise to augment input embedding, we also attempt with adding random uniform noise to input embedding in $\delta$-SAM for fair comparison. 
Following the evaluation setting of Mirror-BERT, we report the average Spearman's $\rho$ under 5 fixed random seeds for all models.
All hyperparameters of FreeLB, R3F, SAM, and $\delta$-SAM are tuned on the development set of STS-B.

\stitle{Results and Discussion.}
Results are shown in~\Cref{tab:unsupervised_sts_full}.
Overall, $\delta$-SAM with random noise achieves the best results on both encoders, outperforming Mirror-BERT$_\textsc{base}$ and Mirror-RoBERTa$_\textsc{base}$ by $1.86\%$ and $1.33\%$ in terms of Spearman's $\rho$ in average, respectively, and also outperforms other methods.
On individual datasets, $\delta$-SAM with random noise achieves the top performance on 11 of 14 setups.
Note that both R3F and FreeLB use random noise so the comparison is fair.
Applying $\delta$-SAM alone leads to improvements of $0.87\%$ and $0.29\%$ on Mirror-BERT$_\textsc{base}$ and Mirror-RoBERTa$_\textsc{base}$, respectively.
Besides, $\delta$-SAM outperforms SAM on both encoders either w/ or w/o random noise.
These results have demonstrated that besides supervised models, $\delta$-SAM can also effectively improve self-supervised sentence embedding models.

\begin{table}[!t]
    \centering
    \scalebox{0.78}{
    \begin{tabular}{p{5.4cm}ccc}
    \toprule
    Model& RG-1& RG-2& RG-L \\
    \midrule
    BART~\cite{lewis-etal-2020-bart}& 44.16& 21.28& 40.90\\
    PEGASUS~\cite{zhang2020pegasus} & 44.17& 21.41& 41.11 \\
    BART + R3F~\cite{aghajanyan2020better}& 44.38& 21.53& 41.17\\
    BART + R-Drop~\cite{liang2021r}& 44.51& \textbf{21.58}& 41.24\\
    BART + $\delta$-SAM& \textbf{44.70}& 21.54& \textbf{41.81} \\
    \bottomrule
    \end{tabular}}
    \caption{Results on CNN/Daily Mail summarization.}
    \label{tab:summ}
\end{table}

\subsection{Summarization}\label{ssec:summ}
\stitle{Task Setup.}
We experiment with the abstractive summarization task on the CNN/DailyMail dataset~\cite{hermann2015teaching}.
Using the large version of BART~\cite{lewis-etal-2020-bart} as the encoder-decoder model, we compare $\delta$-SAM to two regularization methods, including R3F and R-Drop, that have experimented on this task.
Besides, we also compare to PEGASUS~\cite{zhang2020pegasus}, which introduces a self-supervised training objective specifically designed for summarization.
Following the experimental settings of BART, we report metrics including the unigram ROUGE-1 and bigram ROUGE-2 for evaluating the informativeness, and the longest common subsequence ROUGE-L for evaluating the fluency.

\stitle{Results and Discussion.}
Results are shown in~\Cref{tab:summ}.
We observe that $\delta$-SAM achieves the best results in ROUGE-1 and ROUGE-L, outperforming the original BART by $0.54\%$ and $0.91\%$, respectively.
As for ROUGE-2, it also outperforms BART by $0.26\%$ and achieves performance comparable to R3F and R-Drop.
This experiment shows the effectiveness of $\delta$-SAM on optimizing an encoder-decoder model for abstractive summarization.

\subsection{Analysis}\label{ssec:analysis}

The previous experiments have demonstrated that $\delta$-SAM achieves promising improvements on various tasks.
In this section, we analyze whether $\delta$-SAM can derive smaller adversarial risk and how well it approximates per-instance weight perturbation.

We assess the adversarial risks and accuracies of four optimization approaches including vanilla training, SAM, $\delta$-SAM, and per-instance weight perturbation.
We measure the adversarial risk with~\Cref{eq:per_instance} in~\Cref{sec:sam},
which is copied as follows:
\begin{equation*}
    \mathcal{L}_\text{adv} = \frac{1}{N}\sum_{i=1}^N \max_{\bm{\epsilon}_i: \left\lVert\bm{\epsilon}_i\right\rVert_2\le \rho} l_i(\bm{w}+\bm{\epsilon}_i).
\end{equation*}
For all compared methods, we set $\rho=0.05$ in $\mathcal{L}_\text{adv}$.
Due to the high computational cost of per-instance weight perturbation (about 7x of $\delta$-SAM with a batch size of 16), we only conduct experiments on two small datasets: MRPC and RTE.

From the results shown in~\Cref{tab:adv_loss}, 
we observe that $\delta$-SAM achieves smaller adversarial risk than SAM, showing that $\delta$-SAM is indeed a better approximation to per-instance weight perturbation than SAM,
being consistent with our theoretical motivation (\Cref{sec:theoretical_motivation}).
When it comes to accuracy, we observe that:
(1) Per-instance weight perturbation generally achieves the highest accuracy except for the maximum accuracy on RTE, being consistent with the observation in~\citet{foret2020sharpness};
(2) Although $\delta$-SAM consistently outperforms SAM, its performance is often slightly lower than the much more costly per-instance weight perturbation, indicating room for further improvements.

\begin{table}[!t]
    \centering
    \setlength{\tabcolsep}{4pt}
    \scalebox{0.75}{
    \begin{tabular}{lccccc}
    \toprule
    Method& \multicolumn{2}{c}{$\mathcal{L}_\text{adv}$}& \multicolumn{2}{c}{Acc} \\
    &MRPC& RTE& MRPC& RTE \\
    \midrule
    Vanilla & 1.93& 2.97& 84.6/85.3& 67.9/68.2 \\
    SAM& 0.62& 0.78& 86.8/88.2& 68.6/69.3 \\
    $\delta$-SAM& 0.59& 0.75& 87.5/89.5& 69.3/\textbf{72.2} \\
    Per-instance perturbation& \textbf{0.50}& \textbf{0.65}& \textbf{88.2}/\textbf{89.7}& \textbf{70.0}/71.5 \\
    \bottomrule
    \end{tabular}}
    \caption{Adversarial risk and evaluation results on MRPC and RTE datasets. We report the average adversarial risk and the median/max accuracy of 5 runs.}
    \label{tab:adv_loss}
\end{table}



\section{Conclusion}
This paper presents a new sharpness-aware minimization method with dynamic reweighting~($\delta$-SAM).
The proposed method represents the first successful attempt in realizing a \emph{per-instance} weighting scheme. We achieve this by prioritizing instances with larger gradient change rate in adversarial weight perturbation,
in comparison to previous approaches that adopt  \emph{per-batch} weight perturbation.
We show that perturbation calculated on reweighted batch can serve as a better approximation to per-instance weight perturbation while requiring only similar computational cost to per-batch perturbation.
We conduct extensive experiments on the GLUE, STS, and abstractive summarization benchmarks. 
Across all 30 experimental setups that compares to SAM, $\delta$-SAM achieves an consistent improvement over SAM in 27 of them. When compared to a set of other competitive regularization methods, $\delta$-SAM achieves the best performance in 23 out of 33 of the setups.
Further, we quantitatively analyze $\delta$-SAM's impact on sharpness, finding that it indeed leads to flatter loss landscape.
Future work includes inventing new techniques to further reduce the computational cost of $\delta$-SAM and demonstrating its effectiveness on more tasks such as sequence tagging and question answering.

\section*{Limitations}
Like SAM and other training methods based on weight perturbation, the improved performance by $\delta$-SAM is at the cost of introducing additional computational overhead to vanilla training.
Specifically,
although $\delta$-SAM more precisely approximates per-instance weight perturbation with merely $18\%$ extra computational cost to per-batch SAM, 
both SAM and $\delta$-SAM are still slower than vanilla training by roughly doubling the computational costs in practice.
This may limit the application of such optimization algorithms on massive-scale training.

\section*{Acknowledgement}
We appreciate the anonymous reviewers for their insightful comments and suggestions. This material
is supported by the NSF Grant IIS 2105329, a subaward from NSF Cloudbank 1925001 and a Cisco Research Award. 

\bibliography{anthology,custom}
\bibliographystyle{acl_natbib}
\clearpage
\appendix

\begin{center}
    {
    \Large\textbf{Appendices}
    }
\end{center}

\begin{table*}[!t]
    \centering
    \scalebox{0.9}{
    \begin{tabular}{lcccccccc}
    \toprule
        Hyperparameter& MNLI& QQP& RTE& QNLI& MRPC& CoLA&SST2& STS-B\\
        \midrule
        BERT$_\textsc{base}$ \\
        $\rho$& 0.05& 0.05& 0.05& 0.05& 0.05& 0.05& 0.05& 0.01\\
        $\eta$& $\texttt{1e-4}$& $\texttt{1e-4}$& $\texttt{5e-4}$& $\texttt{1e-4}$& $\texttt{1e-4}$& $\texttt{1e-4}$& $\texttt{1e-4}$& $\texttt{1e-4}$\\
        \midrule
        RoBERTa$_\textsc{large}$ \\
        $\rho$&  0.02& 0.02& 0.02& 0.02& 0.02& 0.02& 0.02& 0.01 \\
        $\eta$ &  $\texttt{1e-4}$& $\texttt{1e-4}$& $\texttt{5e-4}$& $\texttt{1e-4}$& $\texttt{1e-4}$& $\texttt{1e-4}$& $\texttt{1e-4}$& $\texttt{1e-4}$\\
        \bottomrule
    \end{tabular}}
    \caption{Hyperparameters for SAM and $\delta$-SAM on the GLUE benchmark.}
    \label{tab:hps}
\end{table*}

\section{Hyperparameters}
On the GLUE benchmark, we search $\rho$ in $\{0.01, 0.02, 0.05\}$ and $\eta$ in $\{\texttt{1e-4}, \texttt{2e-4}, \texttt{5e-4}, \texttt{1e-3}\}$, respectively.
The hyperparameters that achieve the best performance on the GLUE benchmark is listed in~\Cref{tab:hps}.
On the STS benchmark, we search $\rho$ in $\{0.01, 0.02, 0.05, 0.1\}$ and $\eta$ in $\{\texttt{1e-3}, \texttt{1e-2}, 0.1, 1.0\}$, respectively.
We search the scale of uniform random noise in $\{0.01, 0.02, 0.05, 0.1, 0.2, 0.5\}$.
The best hyperparameter of $\rho$, $\eta$, and the scale of uniform random noise is $0.1$, $1.0$, $0.1$ for Mirror-BERT and $0.02$, $1.0$, $0.2$ for Mirror-RoBERTa, respectively.
For summarization, we use $\rho=0.01$ and $\eta=\texttt{1e-4}$.
For all models, we use grid search to find the best hyperparameter.
For other hyperparameters (e.g., batch size, training steps, learning rate, etc.), we directly use the suggested values in the original papers.
Note that for per-instance perturbation, we adopt twice the number of original epochs since we observe they are under trained with default number of epochs.

\section{Running Time}
The running time of SAM and $\delta$-SAM on the GLUE benchmark is shown in~\Cref{tab:time}.\footnote{For easy comparison, the recorded training time is for the same number of epochs across all models, though per-instance perturbation actually takes twice the number of  default epochs in the actual experiment.}
All experiments are conducted on one RTX2080 GPU.
\begin{table*}[!t]
    \centering
    \scalebox{0.9}{
    \begin{tabular}{lcccccccc}
    \toprule
        Running time& MNLI& QQP& RTE& QNLI& MRPC& CoLA&SST2& STS-B\\
        \midrule
        BERT$_\textsc{base}$ \\
        SAM& 1240& 818& 15& 349& 12& 25& 118& 26\\
        $\delta$-SAM& 1350& 995& 16& 450& 15& 25& 132& 31\\
        Per-instance weight perturbation& -& -& 87& -& 105&&&\\
        \midrule
        RoBERTa$_\textsc{large}$ \\
        SAM&  1056& 2591& 33& 1402& 30& 38& 285& 42 \\
        $\delta$-SAM & 1699& 2963& 36& 1425& 35& 45& 325& 58\\
        \bottomrule
    \end{tabular}}
    \caption{Average running time~(in min) for SAM and $\delta$-SAM on the GLUE benchmark.}
    \label{tab:time}
\end{table*}

\end{document}